\documentclass[10pt,twocolumn,letterpaper]{article}

\usepackage{cvpr}
\usepackage{times}
\usepackage{epsfig}
\usepackage{graphicx}
\usepackage{amsmath}
\usepackage{amssymb}

\usepackage[font=footnotesize,labelfont=bf]{caption}

\usepackage{multirow}
\usepackage{tabularx}
\usepackage{dsfont}
\usepackage{url}
\usepackage[tight]{subfigure}
\usepackage{color}
\usepackage{subfigure}
\usepackage{amsthm}
\usepackage[ruled,vlined]{algorithm2e}
\usepackage[export]{adjustbox}

\theoremstyle{definition}

\theoremstyle{remark}

\usepackage[breaklinks=true,bookmarks=false]{hyperref}

\cvprfinalcopy 

\def\cvprPaperID{21} 

\ifcvprfinal\fi
\begin{document}

\title{Class Correlation affects Single Object Localization using Pre-trained ConvNets}
\author{Pokkalla Harsha Vardhan,  Kunal Sekhri, Dipan K. Pal and Marios Savvides\\
Dept. Electrical and Computer Engg.\\
Carnegie Mellon University\\
{\tt\small \{hpokkall, ksekhri, dipanp, marioss\}@cmu.edu}}

\maketitle

\begin{abstract}
 The problem of object localization has become one of the mainstream problems of vision. Most of the algorithms proposed involve the design for the model to be specifically for localizing objects. In this paper, we explore whether a \textbf{pre-trained} canonical ConvNet (\textbf{without} fine-tuning) trained purely for object classification on one dataset with global image level labels can be used to localize objects in images containing a single instance on a separate dataset while generalizing to \textbf{novel} classes. We propose a simple algorithm involving cropping and blackening out regions in the image space called Explicit Image Space based Search (EISS) for locating the most responsive regions in an image in the context of object localization. EISS brings to light the interesting phenomenon of a ConvNets responding more to features within objects as opposed to object level descriptors, as the classes in the training data get more correlated (visually/semantically similar).
\end{abstract}

\section{Introduction}

ConvNets have been used extensively for a variety of different tasks in the vision community. Originally, they were proposed as a technique to address object classification/detection (classifying which object exists in the image) \cite{krizhevsky2012imagenet}. In recent years, it was found that they learn useful representations of objects, and that their features can be used for other (arguably harder) tasks \cite{DBLP:journals/corr/YosinskiCBL14, DBLP:journals/corr/EigenF14, DBLP:journals/corr/SuQLG15, DBLP:journals/corr/FlynnNPS15, DBLP:journals/corr/SimonyanZ14, DBLP:journals/corr/DosovitskiySB14}. These features preserved most of the information regarding the object, however, low level image information (\emph{e.g.} pixel level) was lost which was required for certain tasks like image segmentation leading to emergence of specific architectures and datasets for those specific tasks. Nonetheless, for tasks such as localization, it can be argued that only an intermediate level of information (at the level of super-pixels) is sufficient to perform the task. Nonetheless, most studies, in order to maximize performance, explicitly train architecture and models tuned to the particular task \cite{Girshick_2014_CVPR, DBLP:journals/corr/Girshick15, frrcnn, DBLP:journals/corr/HariharanAGM14a}.  In this paper, we investigate some interesting aspects of the behavior of ConvNets, the representation they learn and of the problem of object localization itself.

\begin{figure}[t]
\centering
\begin{tabular}{cc}
\subfigure[Successful examples]{\includegraphics[scale=.17]{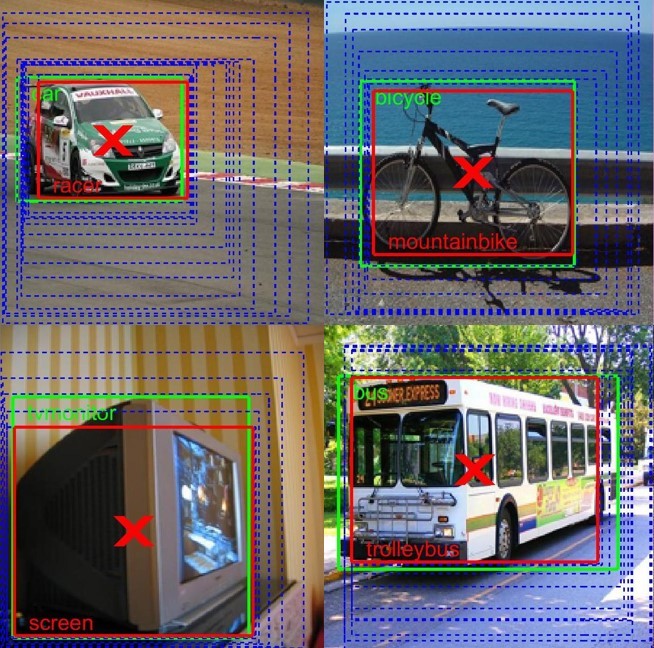}} &
\subfigure[Unsuccessful examples]{\includegraphics[scale=.155]{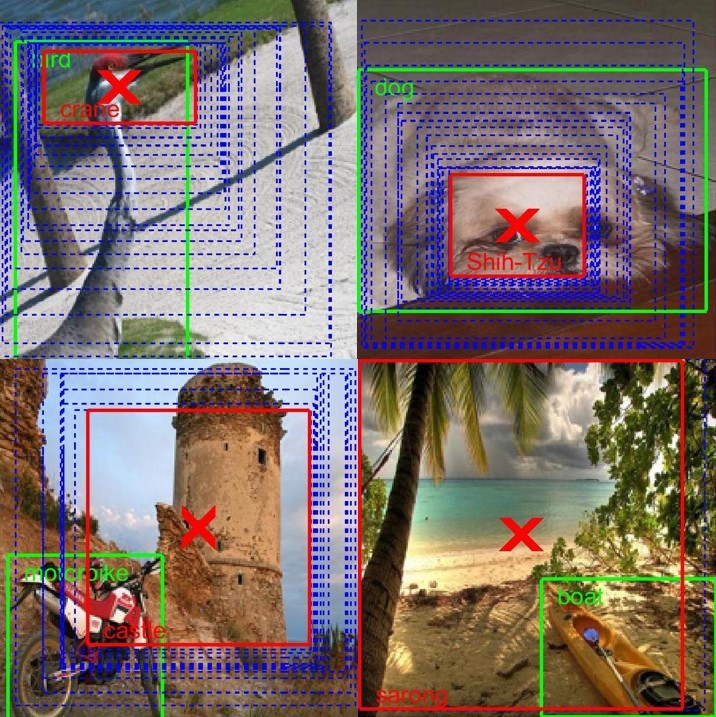}}  
\end{tabular}

\caption{Examples illustrating EISS on AlexNet on samples from the PASCAL 2007 test set. The figure showcases a) successful cases and b) unsuccessful cases. The green box is the ground truth and the red denotes the final output of EISS. The blue boxes show the progression of the algorithm through various iterations. Note the class names. EISS on AlexNet makes interesting kinds of mistakes (showcased in a more detailed figure later). }
\label{fig_setionc_hist}
\end{figure}

\textbf{Motivation.} The most popular algorithm for training ConvNets is supervised. However, although large-scale labeled datasets have started to emerge and gain popularity, it is expensive to obtain such large-scale datasets for finer tasks such as object localization and image segmentation. Thus, the question arises whether it is possible to use only globally labeled images to perform a more challenging task (such as object localization in this paper). This is especially hard since the object can undergo a variety of transformations (scale, translation and rotation) for which the model will have no supervision. Effectively, the model for the harder task is now weakly supervised. In this paper, we focus on the interesting aspects of the problem of addressing the task of object localization (predicting a bounding box, thus the location and extent of an object) using only object classification data (only presence or absence of an object in an image). This would only be possible if the problem of image classification and object localization are inherently coupled, \emph{i.e.} in order to perform well on the image classification task, a model would have to learn a good internal representation that can be used for object localization as well. This is the key observation that helps to address the problem.\\
Further, although deep networks have shown impressive results in a variety of vision tasks, there is a lot of room for improving our understanding of their behavior. Many works have addressed this need in recent years \cite{ZeilerF13, SimonyanVZ13, ZhouKLOT14, NguyenYC14}. This study contributes to this endeavour by investigating blackening out and cropping out regions of the image and studying the change in the response curves. We also highlight some important aspects about the training datasets used to train ConvNets and its implications on the model behavior.

\textbf{Goal.} The goal of this study is not to propose a practical method for object localization, but instead investigating how a \textit{simple} method of iteratively masking and cropping out `interesting' regions in the image can perform on the object localization task. The simplicity of the approach and the lack of fine-tuning also throws light on the intrinsic behavior and biases of the pre-trained image classification model towards the more challenging task. Interestingly, we also find how feature biases can emerge in a model based on the dataset the model is trained on. This phenomenon, in most studies, does not receive the attention it deserves while using features from pre-trained models for various tasks. Hence, investigating this phenomenon helps in a more informed use of pre-trained models in general.

\textbf{ConvNet features can be used for other tasks with additional architecture and training: Model investigated}  The model we focus on heavily in this paper is AlexNet \cite{krizhevsky2012imagenet}  which was  \textbf{pre-trained} on the Imagenet 2012 object classification challenge, and has $\textit{\textbf{not}}$ been fine-tuned in any way. AlexNet has emerged as one of the most popular feature-extractors in the field in recent years. Its use has resulted in a number of studies achieving impressive results in various tasks \cite{Girshick_2014_CVPR}. Even though the model is usually fine-tuned for the particular task, the behavior of the response (features) on a plethora of related inputs (\emph{e.g.} inputs varying in scale) has not been investigated. Given the wide spread use, sometimes even without fine-tuning, it is important we understand this behavior, which arguably help us use the model better.

\textbf{Object Classification and Localization are coupled problems.} The problem of learning a good internal representation has emerged to be one of the core problems in machine learning and vision. ConvNets have brought to light the importance of hierarchy and invariance in a model. Being invariant to common transformations is an important property of any classifier. ConvNets have been understood to be more invariant up the hierarchy, which is good for classification of the entire scene. One could argue that this forces it to lose local level information such as the location and orientation of the object. During training, when only the presence/absence of an object is provided to the model, it needs to learn an internal representation of what the object looks like from across images. However, it can only do so when it knows \textit{where} the object is in order to reinforce its internal representation of the object. Thus, the \textit{problems} of object classification and object localization are coupled, \emph{i.e.} one needs to address \textit{both} sub-problems in order to perform well on either one. Nonetheless, one can try to \textit{decouple} them explicitly. ConvNets address this issue by being invariant (hierarchical pooling) to the transformations the object undergoes. Thus, the localization problem is mitigated with a `dont-care'. In order to leverage ConvNets, we need an inverse map from the label space to the image space. One way to do this would be to explicitly search the image space in a greedy manner for the most informative region affecting the final response. In this paper, this is the approach we adopt.

\textbf{Class Correlation in training data biases ConvNet models.} In our experiments, we make an interesting observation that our search method to find the most responsive regions in the image tends to be more responsive to features regions (regions inside the object boundary) rather than object overall (just outside the object boundary). This phenomenon is explained in more detail in Section~\ref{sec_method}. We find that in order to make ConvNets respond to more global `object' level descriptors, the training data needs to have fewer correlated classes. By correlated classes, we refer to classes whose samples have a high degree of visual similarity for \emph{e.g.} ILSVRC 2012 has a number of classes for dogs, cats and aeroplanes, whereas PASCAL 2007 has a single class for the same. Thus, ILSVRC trains the model to distinguish between different kinds of dogs with the same weight as between a dog and an aeroplane. This biases the model to look for more discriminative features (typically smaller and within the object) rather than object level descriptors which might be similar between different classes (for \emph{e.g.} different kinds of dogs have similar anatomical structure). Hence, using responses as a guide to search might lead to putting boxes around features within the object, thereby reducing the IOU score. 

We verify this phenomenon in our experiments, and conclude that the simple method of blackening does seem to work reasonably well, however, we hypothesize that it would've worked better given training data with less correlation. Knowledge about this phenomenon could be useful for future work where authors need to decide whether they prefer feature (inside object) level descriptors or object (around object) level descriptors for their application. Datasets with less correlation will enable the ConvNets to return descriptors at the object level rather than a sub-part or feature level. Control over correlation in the dataset could be a useful feature in many applications requiring attention to detail (such as image segmentation) or the overall object (such as object localization).


\section{Related Work.} 

The problem of object localization has received a lot of interests over the years since it is one of the fundamental vision challenges \cite{Dalal:2005:HOG:1068507.1069007, Lafferty:2001:CRF:645530.655813, ZhuCYF10, SandeSS14}. It has also been rejuvenated through the use of deep convolutional models with architectures specific to the problem \cite{frrcnn, Girshick_2014_CVPR, overfeat, szegedy2013deep, erhan_local}. However, all of these models (including support/augmented architectures) are designed specifically for the task of object localization and are fine-tuned to maximize performance. Further, the studies mentioned involve manual annotations of local bounding boxes of training data which as datasets get larger, would be more difficult and expensive to obtain. Our simple approach of blackening out the image in order to guide localization of the object does not require such training data, and any pre-trained ConvNet model (and possibly non-neural network models as well) can be utilized \textit{without} the need for fine-tuning.

Masking out or replacing regions in the image space with controlled input has been previously used to analyze the behaviour of ConvNets. An instance of the approach was used to visualize the features learned by the model \cite{ZeilerF13}. In another instance, a similar approach was used to focus on the foreground \cite{DaiH014, hariharan2014simultaneous}. Masking (or blackening in our case) can instead be used to localize the object in addition to focus `attention' to the foreground. Our approach is also related to attention mechanisms for localization  \cite{BaMK14, Cretu2015369} and recognition since our method also employs iterative crops of the image. Thus, the model attends more to a local part of the image in subsequent iterations. Further, blackening out employs attention albeit in a `negative' sense \emph{i.e.} the object localized using information of it not being in the attended region.

Masking out images to perform object localization specifically has also been explored in recent years \cite{BergamoBAT14, oquab2015object}. \cite{oquab2015object} tries to directly answer the question whether global image labels can be leveraged to help with object localization. However, they explore a modified architecture with a specialized training scheme to address the problem. Although this is useful, it is limited in providing deep insights into the behavior of simply pre-trained AlexNet. We therefore, restrict ourselves to the canonical training procedure (standard back-prop with global image level labels) that AlexNet was pre-trained with and perform no further training or fine-tuning of any sort apart from the search algorithm's hyper-parameters.


\section{Explicit Image Space based Search (EISS) for Object Localization}\label{sec_method}

\begin{figure*}
\centering
\includegraphics[width=2\columnwidth]{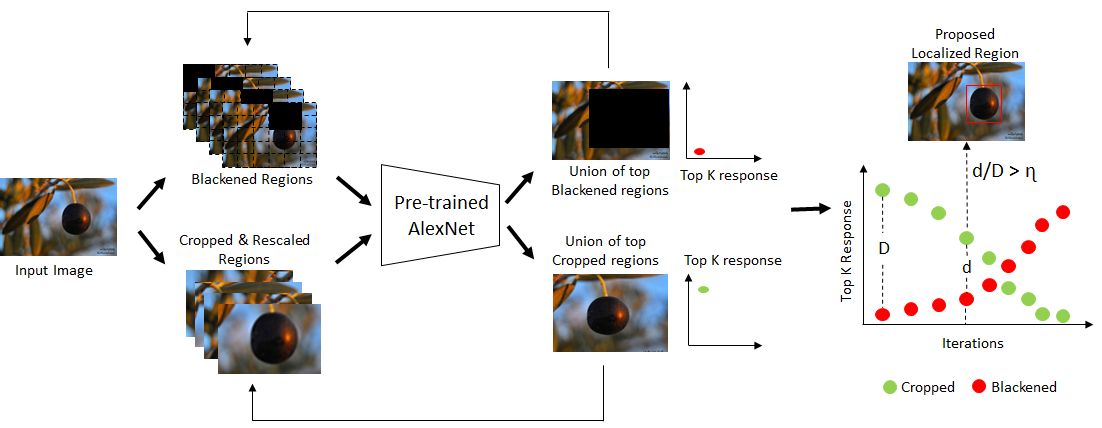}
\centering 
\caption{Explicit Image Space Search}
\label{eiss}
\end{figure*}

The method, called Explicit Image Space based Search (EISS) , we present for localization using pre-trained (without fine-tuning) ConvNets involves an explicit search in the image space. Essentially, the idea is to use the response of the ConvNet to two versions of the image. The first version blacks out a given region in the image (replaces pixels with 0) and the second version crops out the given region and rescales it to allow the ConvNet to compute a response. The chosen parts of the image for the blackening and cropping represent the proposal region for the object. Multiple such regions are proposed and their responses are used to guide the search for the region that the ConvNet responds the most to. We now describe the algorithm in more detail.

\begin{figure*}
\centering
\includegraphics[width=2\columnwidth]{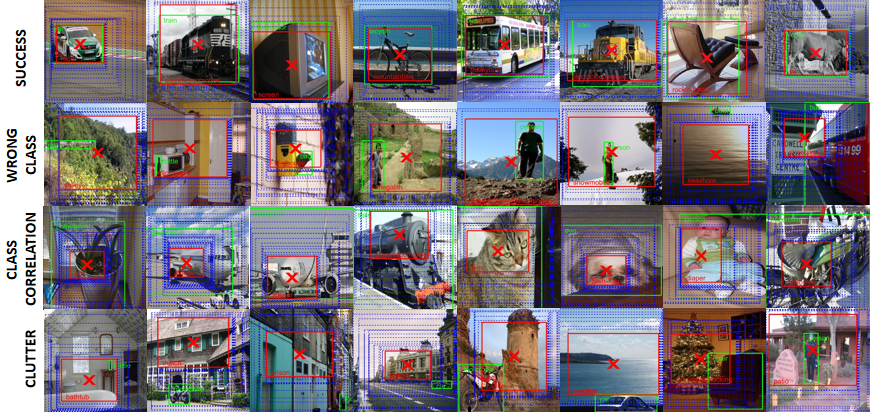}
\centering 
\caption{A few samples showcasing successful cases of localization and failures due to `wrong class', `class-correlation' and `clutter'. The green bounding boxes are the ground truth, red boxes are the final output of EISS over AlexNet and the blue boxes are the iterations EISS went through. The red cross is the location (not including the extent of the object) of the object.}
\label{exp_2_collage}
\end{figure*}

\textbf{Initial response of the model.} Before the actual search begins, the model's response to the original image is saved. The response (the final class probabilities) serve as a reference for the rest of the search. It provides information as to which class features  correlates more with the original image, thereby serving as a heuristic as to which object might exist in the image. The top K classes are then identified. These class identities are the ones whose responses will be used to guide the search through the iterations. 


\textbf{Top-K class response.} We investigate the use of top K classes in the search as opposed to the extreme case of all classes. This is done in order to minimize the diverting effect of the large number of features which are not present in the image. Focusing on the top few classes focuses the search on maximizing the response for a particular object. Alternatively, focusing on just the top class might miss out on useful information for guidance. For instance, a particular image might contain an object which sufficiently fires multiple classes in the model. The other classes in this case could help regularize the search for the object location.

\textbf{Each Iteration.} At every iteration, a set of $N$ regions are proposed which are $0 < \alpha < 1$ times relative in size to the previous iteration. In EISS, we use a stride of 1 resulting in an explicit search over the image for regions $\alpha$ times smaller (random search can improve speed). For every region, two versions of the image are generated, 1) \textit{the blackened version}, wherein the proposed region is replaced by 0's, 2) \textit{the cropped version}, wherein the proposed region is cropped out of the image and re-scaled so as to meet the input specification of a canonical ConvNet model. Thus, the $2N$ modified images (proposals) are passed through the model and the responses (class probabilities) are computed. Then, the top-K class response vector for each proposal is compared with the original global top-K response vector. The proposals (both the blackened and the cropped versions) are then scored using the inner-product between the top-K response vector of each proposal and the original global top-K response vector. Following this, the top 5 scoring regions each from the blackened and the cropped proposals are unioned together (a union of 10 regions in total) to result in one region. \\

\textbf{Blackened and Cropped score.} At the end of every iteration, the resultant region is 1) blackened and 2) cropped out and re-scaled to result in two images. The inner-product between the top-K response vector of each of these images with the global top-K response vector results in two scores, 1) \textit{blackened score} (corresponding to the blackened image at the end of the iteration) and 2) \textit{cropped score} (corresponding to the cropped image at the end of the iteration). Intuition tells us that as the algorithm progresses, with each iteration, typically the blackened score should increase and the cropped score should decrease.

\textbf{Stopping Criterion.} The above procedure can be repeated for many iterations resulting in the 1) \textit{blackened score} typically increasing 2) \textit{cropped score} typically decreasing after a small increase (a phenomenon we investigate more in our experiments). However, they continue to a follow similar trend until the cropped score reaches 0, and the blackened score tends to reach the original score. The proposed region until that time becomes very small, focusing on finer details of the object. Thus, we use the intersection of the curves of the blackened score and the cropped score as a heuristic to stop. The resultant proposed region tends to focus on more of the entirety of the object. We find that this heuristic naturally in most successful cases gravitates towards boxes just \textit{within} the object boundary. In order to respect the ground truth better, we stop a few iterations before the intersection of the blackened and cropped score curves. As our stopping criterion, we stop when difference between the blackened score and the cropped score is less than $\eta\%$ of the initial global top-K response.

\textbf{EISS focuses on features rather than objects.} Since we effectively are searching for which regions drive up the top-K response, regions corresponding to discriminative features between classes would respond the most. Thus, as the algorithm progresses, it will tend to provide \textit{within} the object as opposes to around it. We verify this fact in our experiments. This could be useful in certain applications where we would like to know which part of the image the ConvNet is focusing on for the classification, however for the task of object localization specifically might not be very useful. Early stopping would be one way to address this problem. The stopping parameter $\eta$ can be set so as to stop early enough before the blackened and the cropped curves intersect. 

\textbf{EISS performance on uncorrelated test data can indicate class correlation in training data.} Another interesting aspect of this approach is that it can be used to infer how much class correlation existed in the training data for the model. The mean blackened and cropped response curves for classes that have low correlation (for \emph{e.g.} cat, dog in PASCAL 2007) can indicate whether their existed correlated classes in the training data (for \emph{e.g.} multiple types of cats and dogs in ILSVRC 2012). We verify this phenomenon in our experiments.

\textbf{Random Search for Speed-ups.} EISS can be made significantly faster by randomly selecting $M$ regions instead of explicitly evaluating every region with a stride of 1. For instance, versions of AlexNet which require a batch input size of 32, can take in 16 blackened regions and 16 cropped regions from $M=16$ regions. A sample size of 16 is enough for to gain a fair understanding of the location of the object for reasonably high $\alpha$. Since this paper focuses on the properties and behavior of AlexNet, we do not employ this speed up and use EISS for all our experiments, thus eliminating randomness.

\section{Experiments}

In our experiments, we focus on interesting aspects on application of ConvNets trained purely for classification. For all experiments, unless specified otherwise, we set $\alpha=0.8$, $\eta=10$ and $K=1$.

\subsection{Behavior of Blackened and Cropped Scores  }

\textbf{Motivation.} In our first experiment we investigate the behavior of the blackened and cropped scores curves. Since AlexNet is widely used as a feature extractor for subsequent processing for a variety of tasks, we investigate its characteristics. EISS at subsequent iterations changes the scale of search, which also changes the response of the model. In most applications, this is usually ignored and procedures down the line in the algorithm have to learn to handle it. We explore the change in response in our first experiment.

\textbf{Set-up.} We run EISS on training images from the PASCAL 2012 dataset containing single instances of the image class over 20 different classes for 30 iterations. Note that since we set a max iteration count, we do not need to use $\eta$ in this experiment. Since PASCAL is highly skewed with a few classes have a lot of images and a few having very few, we choose 100 random images for this experiment unless the class contains less than 100 images, in which case we choose all images from that particular class. Thus we report results on \~ 1900 images in total. We compute the mean blackened score and the cropped score curves over all classes, and also for each class. We also run this experiment for $K=5$.

\begin{figure}
\centering
\includegraphics[width=0.8\columnwidth]{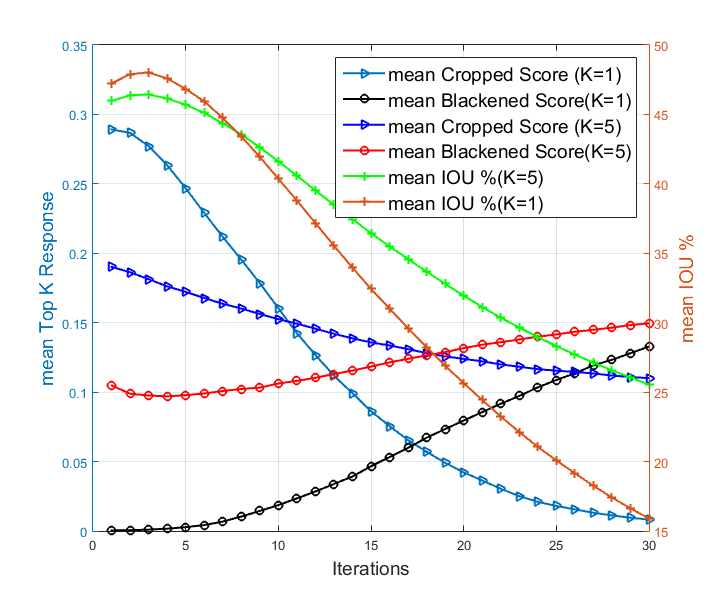}
\centering 
\caption{The mean Top-K response scores and mean IOU \% over all classes for K=1 \& K=5. Response scores and IOU scores are normalized to show the trend.}
\label{exp_1_overall}
\end{figure}

\textbf{Results.} We find that as the algorithm progresses, the blackened score typically increases and the cropped score typically decreases for both $K=1, 5$ as shown in Fig.~\ref{exp_1_overall}. The intersections guarantee that our algorithm converges with a box. We use this plot to also perform parameter tuning for localization. The fact that the IOU decreases after iteration 3, reflects the nature of EISS to focus on discriminative \textit{features} of the object as opposed to the object itself. The discriminative features typically lie \textit{inside} the object and thus the bounding box returned encompasses the discriminative feature instead of the overall box (as shown in a few examples in Fig.\ref{exp_2_collage}). Since the ground truth boxes are slightly larger than the object itself, the IOU decreases as the algorithm progresses. AlexNet was trained to \textit{discriminate} between different classes and thus, discriminative features result in the highest response thereby guiding the search towards that region.

For $K=5$, we find that the blackened curves and the cropped curves vary much less owing to the regularization that additional classes bring in. However, we find that the IOU follows a similar trend although it behaves better as iterations go past 15.


\subsection{Correlated class concepts during training lead to low level feature selectivity}

\textbf{Motivation.} We noted previously that correlated classes in the training data for the ConvNet forces it to respond most to discriminative features between classes. In this experiment, we verify this phenomenon on the training data of PASCAL 2007 containing a single object instance.

\textbf{Set-up.} The set-up is very similar to the previous experiment. We now focus on each PASCAL class separately as opposed to the global behavior of AlexNet. We plot the mean blackened and cropped scores for various classes in PASCAL along with their mean IOU curves.

\begin{figure}[t]
\centering
\begin{tabular}{cc}
\subfigure[Difficult to find discriminative features]{\includegraphics[scale=.3]{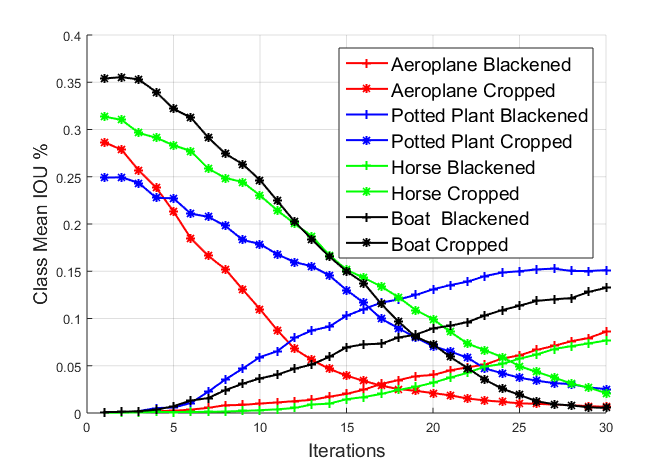}} \\
\subfigure[Easy to find discriminative features]{\includegraphics[scale=.3]{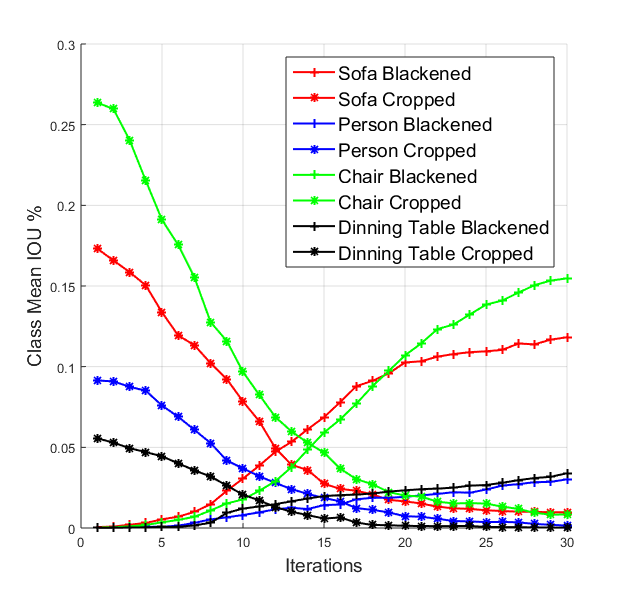}}  
\end{tabular}

\caption{Mean blackened scores and cropped scores for classes difficult/easy to find discriminative features for. EISS on AlexNet takes convergence later/earlier (intersection of blackened and cropped curves). }
\label{exp_2_1}
\end{figure}

\begin{figure}[t]
\centering
\begin{tabular}{cc}
\subfigure[Difficult to localize classes]{\includegraphics[scale=.4]{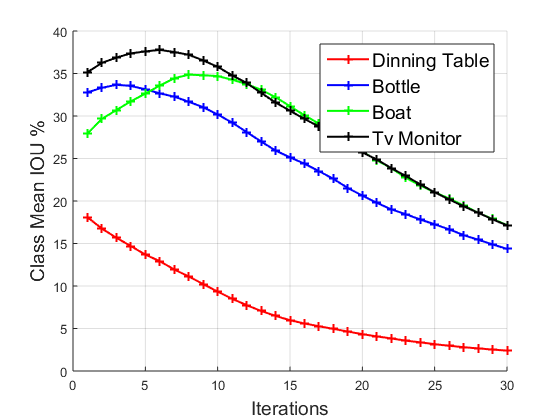}} \\
\subfigure[Easy to localize]{\includegraphics[scale=.4]{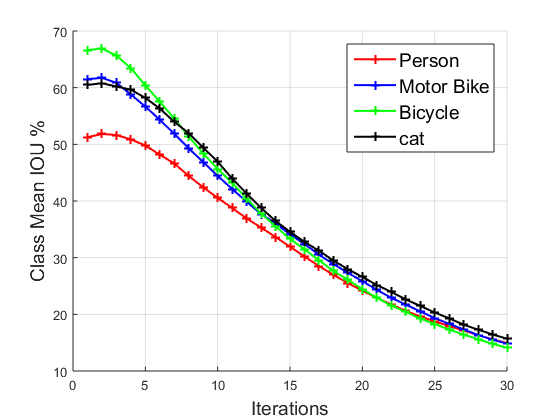}} 
\end{tabular}

\caption{Mean blackened scores and cropped scores for classes difficult/easy to localize. EISS on AlexNet results in lower/higher average IOUs.}
\label{exp_2_2}
\end{figure}

\textbf{Results.} Fig.~\ref{exp_2_1} shows the top 4 classes which we found to be hard/easy to find features in. Whereas, Fig.~\ref{exp_2_2} shows the same for classes which we found hard/easy to localize. Note that even though EISS in many cases finds the object and is able to localize, owing to class-correlation during training, EISS on AlexNet optimizes for a slightly different objective than localization. Depending on the complexity of the class (how many sub parts of the object there exists) and the amount of class-correlation discrepancy in the test and training dataset, localizing the most responding feature regions of the object might correlate with localizing the object itself. 

From Fig.~\ref{exp_2_1} we found that aeroplane, horse, potted plant and boat were the most difficult to find feature or high response regions from (among others were cat, dog, cow). This is sense EISS on average takes more number of iterations to converge. Whereas, dining table, sofa, person and chair were the among the easiest to find high response feature regions. More iterations for convergence implies that AlexNet focuses on lower level features more than object level descriptors. One of the reasons this occurs is class correlation in ILSVRC 2012 which are related to the classes in PASCAL such as  aeroplane, cat and dog. The second reason is that classes such as potted plant tend to have low level discriminative features from other classes (which have visual concepts such as leaves or cylinders) which in turn require more number of iterations to localize. 

Classes such as dining table and sofa have object level discriminative features (due to the absence of correlated classes in ILSVRC 2012 with these PASCAL classes) and thus EISS converges early. The person PASCAL class is an interesting case. A lot of classes in ILSVRC 2012 often involve a person in the extended reaches of the object (such as lollipop, trench-coat), and hence person in PASCAL often have large object level visual concepts which tend to respond high through AlexNet. Note that person is a novel class for AlexNet pre-trained on ILSVRC 2012.

From Fig.~\ref{exp_2_2} we found that dining table, boat, bottle and tvmonitor were the most difficult to localize with respect to the ground truth bounding boxes. Whereas, cat, person, motorbike and bicycle were the among the easiest to find high response feature regions. We found that bottle, and boat were hard to localize due to failure of the EISS to deal with `clutter' and `wrong class'. Failure due to clutter is due to AlexNet responding to other classes in the image which in a completely different region. Failure due to `wrong class' occurs when AlexNet responds to a region close/on the image, however, due to miss-classification it responds to an unrelated local feature and misguides EISS. 

On the other hand, classes such as cat, motorbike and bicycle were easier to localize within the first 15 iterations than the other classes. Thus, setting an $\eta$ to be higher (stopping earlier) boosts the mean IOU ($>50\%$) for all top 4 classes. For a low value of $\eta$ however, classes such as cat saw failures such as `class correlation' as shown in Fig.~\ref{exp_2_collage}. Due to the presence of a large number of samples in ILSVRC 2012 classes related to cat, objects related to person, motorbike and bicycle, these classes seem to be easy to localize in the initial few iterations. 

\textbf{Interesting failure cases for EISS over AlexNet.} Failures can be characterized into three types as illustrated in Fig~\ref{exp_2_collage}. Failures due to `wrong class' occur when the ConvNet recognizes parts of the object as belonging to a different class and thus is misguided through the search. Failures due to `class-correlation' occur when EISS outputs a box focused on a detail or feature \textit{inside} the object rather than around the object. This is due to highly correlated classes in training, leading to the model being sensitive to local discriminative features. Thereby, localization suffers. One method of dealing with this would be early stopping. Failure due to clutter occurs when the image essentially contains instances of multiple objects. AlexNet trained on ILSVRC 2012 expects a single instance. Further, EISS guides the search towards the highest response and thus misses the second object once it leaves the search region.

\section{Conclusion}

We presented a simple method incorporating blackening and cropping out regions in the image space in order to perform localization using a canonical pre-trained ConvNet with no fine-tuning. Our method called EISS employs a sliding window approach since the goal of the study is evaluation rather than proposing a practical method. Randomization can be employed in order to arrive at a more practical algorithm. We find that correlated class concepts in training data result in the model being more selective to low level features rather than object level descriptors. Thus, the EISS algorithm converges slower for such classes. Localization on the other hand, depends on other factors like typical size of the object, correlated classes which have the target class as common (\emph{e.g.} person). We find with just a few iterations of the EISS algorithm, many classes can be localized with sufficient accuracy using a purely pre-trained algorithm with no additional architecture or training.

{\small
\bibliographystyle{ieee}
\bibliography{cvprw_pretrainedCNN}

\begin{thebibliography}{10}\itemsep=-1pt

\bibitem{BaMK14}
J.~Ba, V.~Mnih, and K.~Kavukcuoglu.
\newblock Multiple object recognition with visual attention.
\newblock {\em CoRR}, abs/1412.7755, 2014.

\bibitem{BergamoBAT14}
A.~Bergamo, L.~Bazzani, D.~Anguelov, and L.~Torresani.
\newblock Self-taught object localization with deep networks.
\newblock {\em CoRR}, abs/1409.3964, 2014.

\bibitem{Cretu2015369}
A.-M. Cretu, P.~Payeur, and R.~Laganière.
\newblock An application of a bio-inspired visual attention model for the
  localization of vehicle parts.
\newblock {\em Applied Soft Computing}, 31:369 -- 380, 2015.

\bibitem{DaiH014}
J.~Dai, K.~He, and J.~Sun.
\newblock Convolutional feature masking for joint object and stuff
  segmentation.
\newblock {\em CoRR}, abs/1412.1283, 2014.

\bibitem{Dalal:2005:HOG:1068507.1069007}
N.~Dalal and B.~Triggs.
\newblock Histograms of oriented gradients for human detection.
\newblock In {\em Proceedings of the 2005 IEEE Computer Society Conference on
  Computer Vision and Pattern Recognition (CVPR'05) - Volume 1 - Volume 01},
  CVPR '05, pages 886--893, Washington, DC, USA, 2005. IEEE Computer Society.

\bibitem{DBLP:journals/corr/DosovitskiySB14}
A.~Dosovitskiy, J.~T. Springenberg, and T.~Brox.
\newblock Learning to generate chairs with convolutional neural networks.
\newblock {\em CoRR}, abs/1411.5928, 2014.

\bibitem{DBLP:journals/corr/EigenF14}
D.~Eigen and R.~Fergus.
\newblock Predicting depth, surface normals and semantic labels with a common
  multi-scale convolutional architecture.
\newblock {\em CoRR}, abs/1411.4734, 2014.

\bibitem{erhan_local}
D.~Erhan, C.~Szegedy, A.~Toshev, and D.~Anguelov.
\newblock Scalable object detection using deep neural networks.
\newblock {\em CoRR}, abs/1312.2249, 2013.

\bibitem{DBLP:journals/corr/FlynnNPS15}
J.~Flynn, I.~Neulander, J.~Philbin, and N.~Snavely.
\newblock Deepstereo: Learning to predict new views from the world's imagery.
\newblock {\em CoRR}, abs/1506.06825, 2015.

\bibitem{Girshick_2014_CVPR}
R.~Girshick, J.~Donahue, T.~Darrell, and J.~Malik.
\newblock Rich feature hierarchies for accurate object detection and semantic
  segmentation.
\newblock In {\em The IEEE Conference on Computer Vision and Pattern
  Recognition (CVPR)}, June 2014.

\bibitem{DBLP:journals/corr/Girshick15}
R.~B. Girshick.
\newblock Fast {R-CNN}.
\newblock {\em CoRR}, abs/1504.08083, 2015.

\bibitem{hariharan2014simultaneous}
B.~Hariharan, P.~Arbel{\'a}ez, R.~Girshick, and J.~Malik.
\newblock Simultaneous detection and segmentation.
\newblock In {\em Computer vision--ECCV 2014}, pages 297--312. Springer, 2014.

\bibitem{DBLP:journals/corr/HariharanAGM14a}
B.~Hariharan, P.~A. Arbel{\'{a}}ez, R.~B. Girshick, and J.~Malik.
\newblock Hypercolumns for object segmentation and fine-grained localization.
\newblock {\em CoRR}, abs/1411.5752, 2014.

\bibitem{krizhevsky2012imagenet}
A.~Krizhevsky, I.~Sutskever, and G.~E. Hinton.
\newblock Imagenet classification with deep convolutional neural networks.
\newblock In {\em Advances in neural information processing systems}, pages
  1097--1105, 2012.

\bibitem{Lafferty:2001:CRF:645530.655813}
J.~D. Lafferty, A.~McCallum, and F.~C.~N. Pereira.
\newblock Conditional random fields: Probabilistic models for segmenting and
  labeling sequence data.
\newblock In {\em Proceedings of the Eighteenth International Conference on
  Machine Learning}, ICML '01, pages 282--289, San Francisco, CA, USA, 2001.
  Morgan Kaufmann Publishers Inc.

\bibitem{NguyenYC14}
A.~M. Nguyen, J.~Yosinski, and J.~Clune.
\newblock Deep neural networks are easily fooled: High confidence predictions
  for unrecognizable images.
\newblock {\em CoRR}, abs/1412.1897, 2014.

\bibitem{oquab2015object}
M.~Oquab, L.~Bottou, I.~Laptev, and J.~Sivic.
\newblock Is object localization for free?-weakly-supervised learning with
  convolutional neural networks.
\newblock In {\em Proceedings of the IEEE Conference on Computer Vision and
  Pattern Recognition}, pages 685--694, 2015.

\bibitem{frrcnn}
S.~Ren, K.~He, R.~B. Girshick, and J.~Sun.
\newblock Faster {R-CNN:} towards real-time object detection with region
  proposal networks.
\newblock {\em CoRR}, abs/1506.01497, 2015.

\bibitem{overfeat}
P.~Sermanet, D.~Eigen, X.~Zhang, M.~Mathieu, R.~Fergus, and Y.~LeCun.
\newblock Overfeat: Integrated recognition, localization and detection using
  convolutional networks.
\newblock {\em CoRR}, abs/1312.6229, 2013.

\bibitem{SimonyanVZ13}
K.~Simonyan, A.~Vedaldi, and A.~Zisserman.
\newblock Deep inside convolutional networks: Visualising image classification
  models and saliency maps.
\newblock {\em CoRR}, abs/1312.6034, 2013.

\bibitem{DBLP:journals/corr/SimonyanZ14}
K.~Simonyan and A.~Zisserman.
\newblock Two-stream convolutional networks for action recognition in videos.
\newblock {\em CoRR}, abs/1406.2199, 2014.

\bibitem{DBLP:journals/corr/SuQLG15}
H.~Su, C.~R. Qi, Y.~Li, and L.~J. Guibas.
\newblock Render for {CNN:} viewpoint estimation in images using cnns trained
  with rendered 3d model views.
\newblock {\em CoRR}, abs/1505.05641, 2015.

\bibitem{szegedy2013deep}
C.~Szegedy, A.~Toshev, and D.~Erhan.
\newblock Deep neural networks for object detection.
\newblock In {\em Advances in Neural Information Processing Systems}, pages
  2553--2561, 2013.

\bibitem{SandeSS14}
K.~E.~A. van~de Sande, C.~G.~M. Snoek, and A.~W.~M. Smeulders.
\newblock Fisher and {VLAD} with {FLAIR}.
\newblock In {\em 2014 {IEEE} Conference on Computer Vision and Pattern
  Recognition, {CVPR} 2014, Columbus, OH, USA, June 23-28, 2014}, pages
  2377--2384, 2014.

\bibitem{DBLP:journals/corr/YosinskiCBL14}
J.~Yosinski, J.~Clune, Y.~Bengio, and H.~Lipson.
\newblock How transferable are features in deep neural networks?
\newblock {\em CoRR}, abs/1411.1792, 2014.

\bibitem{ZeilerF13}
M.~D. Zeiler and R.~Fergus.
\newblock Visualizing and understanding convolutional networks.
\newblock {\em CoRR}, abs/1311.2901, 2013.

\bibitem{ZhouKLOT14}
B.~Zhou, A.~Khosla, {\`{A}}.~Lapedriza, A.~Oliva, and A.~Torralba.
\newblock Object detectors emerge in deep scene cnns.
\newblock {\em CoRR}, abs/1412.6856, 2014.

\bibitem{ZhuCYF10}
L.~Zhu, Y.~Chen, A.~L. Yuille, and W.~T. Freeman.
\newblock Latent hierarchical structural learning for object detection.
\newblock In {\em The Twenty-Third {IEEE} Conference on Computer Vision and
  Pattern Recognition, {CVPR} 2010, San Francisco, CA, USA, 13-18 June 2010},
  pages 1062--1069, 2010.

\end{thebibliography}
}

\end{document}